\documentclass[11pt]{article}
\usepackage[preprint]{acl}
\usepackage{times}
\usepackage{latexsym}
\usepackage[T1]{fontenc}
\usepackage[utf8]{inputenc}
\usepackage{microtype}
\usepackage{inconsolata}
\usepackage{graphicx}
\usepackage{amsmath}
\usepackage{amssymb}
\usepackage{booktabs}

\title{{FGSS-BO}: Feedback-Guided Hypothesis Management for LLM-Based Bayesian Optimization}
\author{
  Hao Chen\textsuperscript{1},
  Ziyang Ling\textsuperscript{1},
  Ronald X. Xu\textsuperscript{1,2,*},
  Mingzhai Sun\textsuperscript{1,*}\\
  {\normalsize\textsuperscript{1}School of Biomedical Engineering,
  \textsuperscript{2}School of Engineering Science}\\
  {\normalsize University of Science and Technology of China, Suzhou, China}\\
  {\small\texttt{\{chenhao0873,zaneling\}@mail.ustc.edu.cn}}\\
  {\small\textsuperscript{*}Corresponding authors:
  \texttt{xux@ustc.edu.cn}, \texttt{mingzhai@ustc.edu.cn}}
}

\hypersetup{
  pdftitle={FGSS-BO: Feedback-Guided Hypothesis Management for LLM-Based Bayesian Optimization},
  pdfauthor={Hao Chen, Ziyang Ling, Ronald X. Xu, Mingzhai Sun}
}

\begin{document}
\maketitle
\begin{abstract}
Large language model (LLM)-guided Bayesian optimization (BO) uses past observations to guide search, but proposed directions and their revision can remain implicit. We propose Feedback-Guided Search State (FGSS), a structured hypothesis-management interface connecting language-based guidance to observed evidence, executable parameter interventions, and comparative support updates. An LLM reflector proposes and revises hypotheses; a deterministic curator validates them and maintains identities and test records. The evolving state guides candidate construction and priorities. Instantiated on LLAMBO, FGSS-BO retains its surrogate and default expected improvement selection, with a bounded testing policy. On two retrospective protocol/media tasks with nearest-neighbor response oracles, it attains higher sample mean final best and search AUC than the matched LLAMBO baseline and evaluated non-LLM baselines. Prose-reflection and EI-only controls show mixed component effects. Six direct-training hyperparameter-optimization tasks show task-dependent transfer, strongest on diabetes SVM; exact-grid and reaction controls examine other settings. FGSS exposes hypotheses, their execution, and cross-round revisions for inspection while supporting search gains in selected settings.
\end{abstract}

\section{Introduction}
Bayesian optimization (BO) addresses black-box search when evaluations are expensive, including hyperparameter tuning and scientific experimentation \citep{jones1998efficient,shahriari2016taking,frazier2018tutorial}. Large language models (LLMs) have been incorporated into BO to interpret task descriptions, sample candidates, and predict outcomes from historical examples \citep{liu2024llambo}. These capabilities offer a flexible interface to experimental parameters whose meanings and admissible values differ across tasks.

LLAMBO maintains configuration--outcome observations and conditions its LLM calls on that history \citep{liu2024llambo}. It does not separately maintain identified search hypotheses with evidence references and later test outcomes. This distinction matters under a limited evaluation budget: a suggestion to reduce one parameter may produce a candidate that also changes other coordinates, so its outcome need not directly test the suggested direction. How can an LLM optimizer turn such suggestions into revisable guidance linked to the decisions they influence?

Inspired by reflection and evolving-memory approaches \citep{shinn2023reflexion,suzgun2025dynamic,zhang2025ace}, we propose Feedback-Guided Search State (FGSS) for structured, auditable hypothesis management in BO. Its continually updated guidance consists of search hypotheses with evidence references, an anchor, a parameter intervention, and an expected outcome direction. An LLM reflector proposes and revises these hypotheses; a deterministic curator validates operations and maintains identities and lifecycle records. Accepted hypotheses guide candidate construction and priorities, while eligible matched tests update comparative support for their primary hypothesis. The base surrogate continues to use the observation history.

We instantiate FGSS on LLAMBO and call the resulting optimizer FGSS-BO. LLAMBO separates candidate sampling, surrogate prediction, and acquisition selection, allowing us to introduce hypothesis management into the search loop. FGSS-BO and its matched baseline use the same feasibility-preserving candidate interface, while retaining the LLAMBO surrogate and default EI selection. FGSS-BO additionally permits a bounded exception for hypothesis-test selection.

We assess search performance alongside this hypothesis-management interface. On retrospective iPSC-RPE protocol search and C2C12 culture-media search \citep{kanda2022robotic,cosenza2022multi}, FGSS-BO attains higher sample mean final best and search AUC than its matched baseline. Prose-reflection and testing ablations show mixed component effects. Six preselected hyperparameter-optimization (HPO) tasks use direct model training, with the clearest gains on diabetes SVM; exact-grid, reaction, and chromatography controls examine other settings. Our contributions are:
\begin{itemize}
  \item A structured search-hypothesis representation linking evidence, parameter interventions, and revision and test records through persistent identities.
  \item A search-control mechanism translating LLM-proposed hypotheses into feasible candidate priorities and matched tests through deterministic curation and comparative support updates.
  \item Paired offline protocol/media comparisons with LLM and non-LLM baselines, component ablations, and direct-training HPO transfer checks, supplemented by candidate and oracle diagnostics.
\end{itemize}

\section{Related Work}

\paragraph{Bayesian optimization for scientific discovery.}
BO combines surrogate updating with acquisition-based query selection \citep{jones1998efficient,shahriari2016taking,frazier2018tutorial}. Applications include hyperparameter optimization \citep{snoek2012practical}, robotic iPSC-RPE protocol search, and culture-media development \citep{kanda2022robotic,narayanan2025accelerating}. LLM-guided methods add a language interface to parameter semantics and experimental knowledge within this numerical search loop.

\paragraph{LLM-guided optimization.}
LLM-guided optimizers complement statistical optimization with contextual reasoning and pretrained knowledge. LLAMBO frames BO components in natural language and uses LLMs for warm-starting, candidate sampling, and surrogate prediction \citep{liu2024llambo}; related systems address genetic perturbation design, scientific optimization, and molecular lead optimization \citep{roohani2025biodiscoveryagent,yang2025reasoningbo,zhang2025autolead}. \citet{gupta2025llmsbo} use randomized-feedback controls to diagnose outcome sensitivity in BioDiscoveryAgent. Whereas these controls measure dependence on feedback content, FGSS addresses how an optimizer represents and revises search hypotheses and turns them into search actions.

\paragraph{Feedback, memory, and reflection in LLM agents.}
Self-Refine iteratively revises outputs using self-feedback, while Reflexion retains reflective experience across trials \citep{madaan2023selfrefine,shinn2023reflexion}. Dynamic Cheatsheet accumulates reusable guidance, and Agentic Context Engineering (ACE) organizes evolving context through reflection and curation \citep{suzgun2025dynamic,zhang2025ace}. FGSS adopts these memory and curation principles for numerical BO. Its hypothesis records specify an observed anchor, a parameter intervention, and an expected outcome direction, making the guidance executable. Candidate lineage connects eligible numerical comparisons to the corresponding hypothesis, while the surrogate retains the observation history. Our prose-summary control tests a simpler continuously updated memory interface; FGSS additionally maintains individual hypotheses and their test records.

\section{Background and Problem Setup}

\paragraph{Bayesian optimization and LLM-guided optimization.}
We consider maximization of an expensive black-box objective $f:\mathcal{X}\rightarrow\mathbb{R}$ under a limited evaluation budget. Starting from $K$ initial observations $\mathcal D_0$, after $t$ subsequent search evaluations the history is
\begin{equation}
\mathcal{D}_t=\mathcal D_0\cup\{(x_i,y_i)\}_{i=1}^{t}, \quad y_i=f(x_i).
\end{equation}
BO then selects an unevaluated configuration by maximizing an acquisition criterion over the remaining search space. LLAMBO augments this loop with LLM-based candidate sampling and surrogate prediction while conditioning prompts on task descriptions and previous evaluations \citep{liu2024llambo}.

\paragraph{Explicit search-hypothesis management.}
We augment $\mathcal D_t$ with a state $S_t$ of search hypotheses, evidence references, revisions, and eligible test outcomes. Unlike the observations themselves, these records specify proposed actions whose guidance can change across rounds.

\paragraph{Optimization settings.}
The method accepts numerical, ordinal, and categorical variables. The primary retrospective tasks use a historical-table oracle $\widehat f_{\mathrm{NN}}$, not new laboratory measurements. Exact-grid, emulator/simulator, and direct-training HPO tasks test other interfaces. We present the method for maximization; minimization reverses outcome comparisons and uses the corresponding improvement criterion.

\section{Method}

\subsection{Framework Overview}
\begin{figure*}[t]
  \centering
  \includegraphics[width=\textwidth]{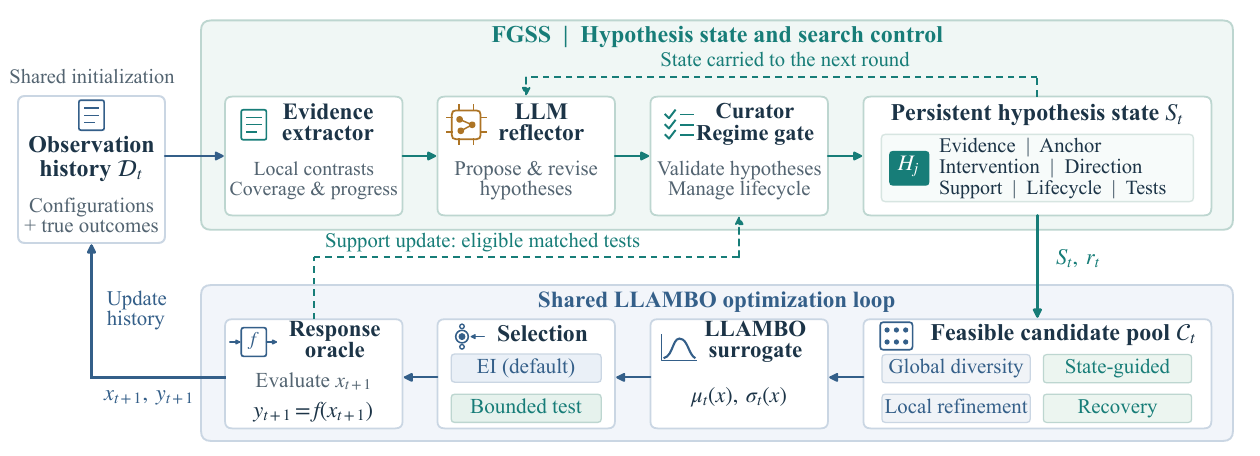}
  \caption{Overview of FGSS-BO. The extractor, LLM reflector, and deterministic curator maintain evidence-linked hypotheses; a regime gate controls candidate allocation. LLAMBO retains its history-conditioned surrogate and default EI selection, with a bounded exception for eligible matched tests. Solid arrows show current-round flow; dashed arrows show cross-round memory and eligible comparative support updates.}
  \label{fig:fgss-overview}
\end{figure*}

Figure~\ref{fig:fgss-overview} summarizes the loop: evidence $e_t$ is extracted from $\mathcal D_t$, the reflector and curator revise $S_t$, and the state guides candidate set $\mathcal C_t$. For example, a hypothesis may propose reducing one parameter at an observed anchor. An exact intervention changes that parameter alone; comparison with the anchor updates the parent hypothesis, informing later retention, revision, or retirement. Appendix~\ref{sec:reflector-state} provides the prompt contract and a recorded state example. The active set is bounded, with archived and retired records retained.

\subsection{Evidence and Feedback State}
The extractor converts recent and global observations into a compact evidence object,
\begin{equation}
e_t=E(\mathcal{D}_t).
\end{equation}
It contains ranked high- and low-performing configurations, recent improvement, stagnation, per-parameter coverage, duplicate regions, and local contrasts between nearby configurations with different outcomes. Local contrasts provide an observed basis for proposing or revising hypotheses.

At round $t$, $S_t^-$ is the carried state after $t$ search evaluations, with $S_0^-=\varnothing$; $S_t$ is its curated revision used for selection. The reflector and curator update it using $e_t$:
\begin{equation}
\begin{aligned}
h_t &= R(e_t,S_t^-), \\
S_t &= C(S_t^-,h_t,e_t).
\end{aligned}
\end{equation}
Each hypothesis contains a persistent identifier, parameter scope, anchor configuration or region, intervention, expected outcome direction, evidence references, uncertainty, a heuristic support score, and lifecycle status. Direct tests target one or two variables. The intervention specifies a testable prediction, and its recorded lineage links a later observation to the primary hypothesis. The curator validates additions and revisions, merges duplicates, manages lifecycle, and caps the active set. For main-effect and interaction hypotheses, support and lifecycle updates retain the ID; changed interventions require new cards.

\subsection{Comparative Support and State Evolution}
Each candidate records the hypotheses contributing to its construction or priority. Let $J_t(x)$ index these contributors within the active state. This lineage is distinct from direct-test attribution:
\begin{equation}
A_t(x)=\{H_j:j\in J_t(x)\}.
\end{equation}
Only an eligible direct test updates its primary hypothesis, with $m_{j,t}\in\{0,1\}$ and $\sum_j m_{j,t}\leq1$. Its evidence weight $w_{j,x}\in[0,1]$ discounts uncertainty, imperfect matching, and control reuse. For control outcome $y^{\mathrm{ctrl}}_{j,t}$, define, for maximization,
\begin{equation}
\begin{aligned}
\delta_{j,t} &= y_{t+1}-y^{\mathrm{ctrl}}_{j,t}, \\
u_{j,t} &= \eta_+\mathbf{1}[\delta_{j,t}>\epsilon_{j,t}] \\
&\quad-\eta_-\mathbf{1}[\delta_{j,t}<{-\epsilon_{j,t}}], \\
s_{j,t} &= m_{j,t}w_{j,x_{t+1}}u_{j,t}
, \\
q_{j,t+1} &= \operatorname{clip}(\bar q_{j,t}+s_{j,t},0,1).
\end{aligned}
\end{equation}
Here $\bar q_{j,t}\in[0,1]$ is the curated heuristic support score in $S_t$, including initialization of new cards; it is not a calibrated probability. Post-evaluation records, including $q_{j,t+1}$, enter $S_{t+1}^-$. The increments $\eta_+,\eta_-$ are fixed, and $\epsilon_{j,t}$ combines an empirical effect threshold with a numerical floor, not a significance test. Staleness subtracts $\lambda\geq0$ once per verification epoch, separately from outcome updates. Minimization reverses $\delta_{j,t}$. Invalid matches receive no outcome update; global-incumbent progress is tracked separately. Open-space tests hold non-target variables fixed; table-backed tests allow bounded nuisance differences and receive discounted weight. Thus $\delta_{j,t}$ supplies comparative support under the oracle, not an identified experimental causal effect. Test counts, control reuse, and failures govern subsequent weakening or retirement (Appendix~\ref{sec:algorithm-details}).

\subsection{Regime Control and Candidate Generation}
The reflector proposes an exploration--exploitation regime, while a deterministic controller constrains the final decision:
\begin{equation}
\begin{aligned}
\widetilde{r}_t &= R_{\mathrm{regime}}(e_t,S_t^-), \\
r_t &= G(\widetilde{r}_t,b_t,g_t,c_t).
\end{aligned}
\end{equation}
Here $b_t$ is the current best outcome, $g_t$ describes recent progress or stagnation, and $c_t$ summarizes coverage. The controller constrains pure exploitation, preserves diversity, and triggers recovery under repeated low-yield search. During recovery, FGSS-BO proposes evidence-linked alternatives to failed directions, for example an interaction hypothesis under which a single-parameter effect depends on another parameter.

For typed parameter spaces, FGSS-BO samples feasible, previously unevaluated configurations and biases their priority using $S_t$ and $r_t$:
\begin{equation}
\mathcal{C}_t=G_{\mathrm{cand}}(
\mathcal{X}\setminus\mathcal{X}_t;S_t,r_t).
\end{equation}
Here $\mathcal X_t$ is the set of configurations evaluated in $\mathcal D_t$. Domain specifications enforce variable types and admissible ranges. The generator allocates candidates among global diversity, local refinement, hypothesis testing near anchors, and recovery. Hypothesis scopes, directions, and anchor contexts guide construction and priority; the global branch preserves proposals outside active hypotheses.

\subsection{Integration with LLAMBO}
FGSS-BO retains the LLAMBO surrogate and its expected improvement criterion. With LLM-predicted mean $\mu_t(x)$ and standard deviation $\sigma_t(x)$ floored at $10^{-5}$ in the inherited implementation, EI over $\mathcal{C}_t$ is
\begin{align}
\mathrm{EI}_t(x)
  &=\Delta_t(x)\Phi(z_t)+\sigma_t(x)\phi(z_t), \\
z_t&=\frac{\Delta_t(x)}{\sigma_t(x)},
\end{align}
where $\Phi$ and $\phi$ are the standard normal CDF and density, and $\Delta_t(x)=\mu_t(x)-b_t$ for maximization. Let $\mathcal C_t^{\mathrm{test}}\subseteq\mathcal C_t$ contain eligible matched interventions, and define
\[
\begin{aligned}
x_{t+1}^{\mathrm{EI}}
&=\arg\max_{x\in\mathcal{C}_t}\mathrm{EI}_t(x),\\
x_{t+1}^{\mathrm{test}}
&=\arg\max_{x\in\mathcal{C}^{\mathrm{test}}_t}P_t(x).
\end{aligned}
\]
The latter is the highest-priority eligible matched intervention. The deterministic priority $P_t$ is lexicographic: adaptive test priority, EI, support-weighted match quality, challenger status, and distance/index tie-breaks (Appendix~\ref{sec:algorithm-details}). FGSS-BO selects
\begin{equation}
x_{t+1}=
\begin{cases}
x_{t+1}^{\mathrm{test}}, & a_t=1,\\
x_{t+1}^{\mathrm{EI}}, & a_t=0,
\end{cases}
\end{equation}
where $a_t$ can be one only when an eligible matched test exists and the testing budget and minimum interval permit it. For $T=20$, the policy has four base test slots and up to two additional challenger slots during stagnation, with a minimum interval of three rounds. These constrain policy overrides, not all direct tests: EI can also select a test candidate, which consumes base-test usage. The bounded exception supplies comparisons linked to specific hypotheses; realized overrides and match quality are reported in Appendix~\ref{sec:test-policy-audit}.

The matched LLAMBO baseline shares the candidate interface and surrogate but omits state conditioning and test-policy selection. Experimental Setup specifies the remaining shared settings.

\section{Experimental Setup}
\label{sec:experimental-setup}

\subsection{Datasets and Tasks}
The primary protocol/media tasks expose configuration--outcome feedback under limited budgets with different variable types.

\paragraph{iPSC-RPE.}
The task concerns differentiation of induced pluripotent stem cells into retinal pigment epithelial cells, maximizing pigmentation score \citep{kanda2022robotic}. We use the \texttt{253G1\_results\_r5r7r10} table. The seven settings cover preconditioning concentration and duration, detachment duration and pipetting, and differentiation supplement schedules. Ordered finite concentration/detachment levels, integer durations and speed, and a binary pipetting-length setting define a discrete configuration space.

\paragraph{C2C12.}
We use a 209-row processed table from the upstream DBO archive associated with multi-information-source culture-media optimization \citep{cosenza2022multi}. Fourteen media coordinates are continuous on $[0,1]$ in the supplied representation. All methods maximize the archived response score on its supplied scale, without additional standardization. The \texttt{source} query label selects among four measurement sources for response lookup rather than specifying a physical culture intervention. Media and source-evaluation costs are outside the objective. Appendix~\ref{sec:c2c12-provenance} details the table provenance and response interpretation.

\paragraph{Response and initialization protocol.}
Both tasks return the score of a nearest historical configuration in normalized parameter coordinates; C2C12 matching is restricted by source. This oracle permits new parameter combinations but does not validate their physical response. Initialization uses the ten lowest-scoring historical observations, shared across methods, followed by twenty search evaluations. This outcome-conditioned low-start protocol is a retrospective stress test, not label-free prospective initialization. The exact-grid iPSC control restricts candidates to observed configurations and removes nearest-neighbor assignment.

Cross-domain checks include six preselected Bayesmark HPO tasks from the LLAMBO repository \citep{liu2024llambo}: breast, digits, and diabetes, each paired with RandomForest and SVM. These train models directly without a nearest-neighbor oracle. Classification maximizes five-fold CV accuracy; diabetes minimizes positive CV MSE. Initialization uses ten label-free public configurations, shared within each pair, followed by twenty search evaluations. Held-out test scores never enter optimization feedback (Appendix~\ref{sec:hpo-transfer}). We also retain all five Summit reaction tasks \citep{felton2021summit,hone2017kinetic,reizman2016suzuki} and Olympus HPLC \citep{hase2021olympus} in the appendix.

\subsection{Baselines}
We evaluate a feasibility-preserving adaptation of LLAMBO (\emph{LLAMBO-Feasible}, hereafter \textbf{LLAMBO}). It retains the LLAMBO surrogate and EI, with the shared typed sampling/refinement interface but no FGSS state or test-policy overrides. The free-form variant, \textbf{LLAMBO-Freeform}, is evaluated separately for candidate-interface diagnostics; it is an implementation-level variant rather than a strict reproduction of original LLAMBO.

We also compare \textbf{Random Search} over the legal domains, \textbf{HEBO} \citep{cowenrivers2022hebo}, and \textbf{GP-EI} implemented in BoTorch \citep{balandat2020botorch}. GP-EI standardizes outcomes and normalizes numerical coordinates, using categorical kernels for pipetting length or measurement source. It optimizes EI over legal discrete settings for iPSC and continuous coordinates with enumerated sources for C2C12. All five methods share the actual initial observations, admissible domains, evaluation budget, and response oracle. Baseline fitting and acquisition settings are detailed in Appendix~\ref{sec:nonllm-details}.

The iPSC comparison adds \textbf{Reflective Summary}: each round, the same extractor's evidence and previous prose summary are used to produce a replacement summary for the shared candidate refiner. It has no hypothesis curation or tests. \textbf{FGSS w/o Tests} retains structured reflection and ordinary state guidance but disables matched-test generation, test-policy overrides, and direct support updates. \textbf{No-Override} retains test candidates and support updates but always selects by EI. Reflection settings and evaluation budgets are shared, not realized token totals. HPO compares the two complete LLM pipelines.

\subsection{Metrics}
We report final best-observed objective value and search-stage best-so-far area under the curve (AUC). Let $K$ be the initialization count and $T$ the subsequent search budget. For maximization,
\begin{equation}
\begin{aligned}
B_t &= \max_{(x,y)\in\mathcal D_t} y, \\
\mathrm{AUC}_{\mathrm{search}} &= \frac{1}{T}\sum_{t=1}^{T}B_t.
\end{aligned}
\end{equation}
Final best is $B_T$; for minimization, replace $\max$ by $\min$. Search AUC averages attained performance after initialization, not ROC AUC. Tables report means and sample standard deviations over six paired runs; bands are standard deviations. Paired bootstrap intervals are exploratory and unadjusted for multiple comparisons (Appendices~\ref{sec:paired-differences}, \ref{sec:component-details}, and \ref{sec:hpo-transfer}). Metrics always use true oracle outcomes. Accuracy and MSE are analyzed separately.

\subsection{Implementation Details}
The primary tasks use $K=10$, $T=20$, and batch size one. The shared LLAMBO configuration uses GPT-4o mini, ten candidates, two templates, and ten generations. We use this backend for its practical balance of instruction following and API cost. FGSS adds a reflection call each round at temperature zero with an output cap of 1500 tokens and one configured retry, increasing total LLM compute. Table~\ref{tab:reflector-usage} reports its recorded per-run token use and elapsed time. Regime cutoffs use supplied task-response units (Table~\ref{tab:reliability-cutoffs}). Optional domain-prior cards and injections are disabled; task descriptions and pretrained LLM knowledge remain available. The appendix specifies settings, prompts, state examples, and provenance.

\section{Results}

\subsection{Main Results on Protocol and Media Search}
Table~\ref{tab:main-results} compares five methods under the shared low-start, nearest-neighbor protocol. FGSS-BO has the highest sample mean final best and search AUC in both primary tasks. Relative to LLAMBO, the mean differences are $+0.123/+0.147$ for iPSC final best/AUC and $+0.445/+0.141$ for C2C12.

On iPSC, final-best margins over GP-EI and Random Search are modest ($+0.020$ and $+0.033$), while the AUC margin over GP-EI is $+0.135$. The paired iPSC interval versus GP-EI includes zero for final best but lies above zero for AUC (Table~\ref{tab:paired-differences}). C2C12 intervals versus GP-EI cross zero on both metrics, as does its AUC interval versus LLAMBO.

\begin{table*}[t]
  \centering
  \footnotesize
  \setlength{\tabcolsep}{3pt}
  \begin{tabular}{llccccc}
    \toprule
    \textbf{Task} & \textbf{Metric} & \shortstack{\textbf{Random}\\\textbf{Search}} & \textbf{HEBO} & \textbf{GP-EI} & \textbf{LLAMBO} & \textbf{FGSS-BO} \\
    \midrule
    iPSC-RPE & Final best & $0.780\pm0.078$ & $0.569\pm0.169$ & $0.793\pm0.026$ & $0.691\pm0.117$ & $0.813\pm0.023$ \\
    iPSC-RPE & Search AUC & $0.557\pm0.150$ & $0.465\pm0.100$ & $0.505\pm0.145$ & $0.494\pm0.057$ & $0.640\pm0.117$ \\
    \midrule
    C2C12 & Final best & $0.790\pm0.340$ & $1.026\pm0.623$ & $1.516\pm0.904$ & $1.281\pm0.220$ & $1.725\pm0.411$ \\
    C2C12 & Search AUC & $0.552\pm0.289$ & $0.722\pm0.404$ & $0.868\pm0.653$ & $1.025\pm0.264$ & $1.167\pm0.256$ \\
    \bottomrule
  \end{tabular}
  \caption{Primary offline protocol/media search over six paired runs: mean $\pm$ sample standard deviation, higher is better. All methods share ten low-score initial observations and twenty search evaluations under the same task-specific nearest-neighbor response oracle. LLAMBO uses the shared feasibility-preserving interface.}
  \label{tab:main-results}
\end{table*}

\begin{figure*}[t]
  \centering
  \includegraphics[width=\textwidth]{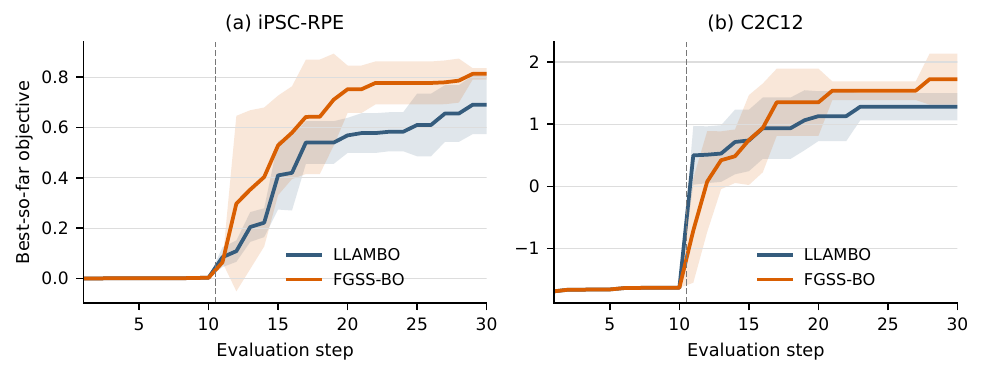}
  \caption{Primary-task mean best-so-far objective over six paired runs. The dashed line separates ten shared low-score initialization observations from twenty search evaluations; shading shows one sample standard deviation. LLAMBO and FGSS-BO share the feasible-candidate interface and nearest-neighbor oracle. C2C12 scores refer to the retained configuration--source query task.}
  \label{fig:main-curves}
\end{figure*}

Figure~\ref{fig:main-curves} shows the paired LLM trajectories; Figure~\ref{fig:biological-fiveway} compares all five methods (Appendix~\ref{sec:nonllm-details}).

\subsection{Component Ablations}
Table~\ref{tab:component-ablations} places prose reflection and EI-only selection alongside full FGSS. Reflective Summary already raises both sample means relative to LLAMBO. Full FGSS exceeds it by $0.0285$ in final best but is $0.0061$ lower in search AUC. No-Override also retains higher means than LLAMBO: its AUC exceeds full FGSS's, while final best is slightly lower, with both paired intervals crossing zero (Figure~\ref{fig:no-override}). Against FGSS w/o Tests, full FGSS's differences are $+0.0088/-0.0275$, again with intervals crossing zero (Appendix~\ref{sec:component-details}). These controls show mixed effects on objective progress despite the differences in hypothesis management. Figure~\ref{fig:immediate-cost} reports the immediate opportunity cost of full FGSS's 18 selection changes (Appendix~\ref{sec:test-selection-effects}).

State-only feedback permutation gives mixed effects: aligned feedback yields a higher final-best mean and lower final-result variability, whereas permutation yields a higher search AUC. Both paired difference intervals include zero (Appendix~\ref{sec:state-feedback-permutation}).

\begin{table}[t]
\centering
\small
\begin{tabular}{lcc}
\toprule
Method & Final best & Search AUC\\
\midrule
LLAMBO & $.691\pm.117$ & $.494\pm.057$\\
Reflective Summary & $.785\pm.061$ & $.646\pm.054$\\
FGSS w/o Tests & $.804\pm.070$ & $.668\pm.051$\\
FGSS No-Override & $.802\pm.110$ & $.690\pm.098$\\
Full FGSS & $.813\pm.023$ & $.640\pm.117$\\
\bottomrule
\end{tabular}
\caption{iPSC component comparison, six paired runs; mean $\pm$ sample standard deviation. All variants share the low-start nearest-neighbor protocol. No-Override retains test candidates and support updates but always selects by EI. LLAMBO uses the shared feasibility-preserving interface.}
\label{tab:component-ablations}
\end{table}

\subsection{Direct-Training HPO Transfer}
Tables~\ref{tab:hpo-transfer} and~\ref{tab:hpo-differences} report the complete six-task HPO results and paired differences. Diabetes SVM provides the clearest gain: mean final CV MSE falls from $0.57768$ to $0.51711$ ($10.48\%$), and search AUC from $0.61046$ to $0.55636$ ($8.86\%$). Breast SVM improves modestly; breast RF and digits SVM are nearly tied. Digits RF and diabetes RF have mixed final/AUC results. These direct evaluations show task-dependent transfer beyond the table-backed oracle.

\subsection{Candidate-Sampling Diagnostics}
\begin{figure*}[t]
  \centering
  \includegraphics[width=0.86\textwidth]{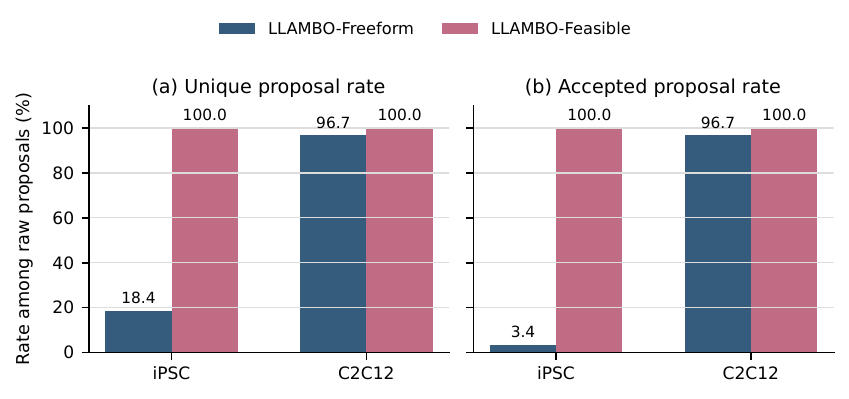}
  \caption{Proposal-level candidate diagnostics from earlier runs, comparing free-form LLM proposals with feasibility-preserving construction. Severe iPSC-RPE duplication motivates a shared feasible-candidate interface.}
  \label{fig:candidate-diagnostics}
\end{figure*}

Figure~\ref{fig:candidate-diagnostics} diagnoses earlier free-form duplication and rejection. On iPSC, 18.4\% of raw proposals are unique and 3.4\% survive validity/novelty checks, motivating legal-space construction. The six-seed interface controls in Figure~\ref{fig:freeform-feasible} show that Freeform remains viable on C2C12, with sample means close to Feasible (Appendix~\ref{sec:freeform-interface}). Both main LLM methods use the same feasible interface across tasks, holding candidate-interface handling fixed in the FGSS comparison.

\subsection{Oracle and Cross-Domain Controls}
Exact-grid iPSC restricts candidates to observed configurations and removes nearest-neighbor assignment. Final-best means are close ($0.840$ LLAMBO, $0.848$ FGSS-BO), while mean search AUC favors LLAMBO ($0.750$ versus $0.710$; Appendix~\ref{sec:exact-grid}). This pattern suggests protocol-dependent performance; the control changes both the candidate domain and oracle assignment. On all five Summit tasks, HEBO and GP-EI outperform the two LLM methods in sample mean final best and AUC; FGSS-BO exceeds LLAMBO on four final-best means and three AUC means, but not Reizman Case 4 (Appendix~\ref{sec:summit-cross-domain}). HPLC yields higher FGSS-BO means on both metrics (Appendix~\ref{sec:hplc-control}).

\section{Discussion}
FGSS organizes hypothesis management around inspectable records: a proposed direction identifies parameters, an anchor, and an intervention whose later comparison can revise the same record. The LLM interprets evidence and proposes changes; fixed rules govern admissibility and numerical support updates. Prose reflection and EI-only selection also achieve higher sample means than LLAMBO, so the component results separate this record-and-action interface from a uniform optimization advantage for full FGSS.

A candidate can outperform its matched control without improving the incumbent. Local comparative support and global progress therefore answer different questions, and FGSS tracks them separately. The higher AUC of No-Override and FGSS w/o Tests suggests a trade-off between early objective progress and obtaining targeted comparisons under this protocol.

HPO extends the evaluation to directly trained objectives, with the strongest gains on diabetes SVM. Exact-grid iPSC's lower mean FGSS AUC shows that state-guided priorities need not improve progress in a restricted candidate domain. Together with the reaction controls, this motivates adapting guidance and testing to the search interface and response landscape. Future refinements could preserve more global exploration when local hypotheses dominate and use observed test utility to allocate the testing budget.

\section{Conclusion}
\label{sec:conclusion}
FGSS provides structured hypothesis management for LLM-guided BO, turning proposed directions into executable parameter interventions and linking eligible comparisons to later revisions. FGSS-BO retains LLAMBO's surrogate and default EI selection, with a bounded testing policy. Its records expose the path from language-based hypotheses to search actions and updates. The complete system achieves higher sample means on the two retrospective protocol/media tasks and yields gains on some direct-training HPO tasks, while component and oracle controls show mixed effects across settings.

\section{Limitations}
\label{sec:limitations}
The primary tasks use historical-table nearest-neighbor responses and outcome-conditioned low-score initialization. Their results describe that retrospective protocol, not label-free prospective laboratory search. C2C12 retains a measurement-source variable, so the query score should not be interpreted as an assay-independent biological optimum. Prospective validation is needed to assess measurement noise, batch effects, experimental failures, and temporal drift.

Evaluation uses one LLM backend and base optimizer; FGSS adds reflection cost. The HPO suite compares the two complete LLM pipelines on directly trained objectives. Future evaluation could extend paired runs to ten seeds and examine further backends, initialization policies, budgets, and optimizers; noise-aware support updates and test prioritization could refine the balance of local hypotheses and exploration.

\bibliography{references}

\appendix
\section{Additional Experimental Details}
\label{sec:appendix}
Algorithm settings (Appendix~\ref{sec:algorithm-details}) and prompt/state examples (Appendix~\ref{sec:reflector-state}) precede the detailed experimental controls below.
\subsection{Algorithm Details}
\label{sec:algorithm-details}
The following specifies the parameterized, sequential full-FGSS implementation. Scientific-task objectives are maximized; diabetes HPO minimizes positive MSE, reversing improvement and support-update signs. $S$ supplements the observation history with identified hypotheses, revisions, and test records, including persistent archived/retired memory; only the active set is capped. The reflector receives compact cards for all three groups, so its input can grow with retained memory. Candidate guidance includes at most four active cards.

\paragraph{Full FGSS loop.}
Input: typed search space $\mathcal X$, oracle $f$, paired initial observations $\mathcal D_0$, search budget $T$, candidate count $N$, and empty carried state $S_0^-$.
\begin{enumerate}
  \item For each search round $t=0,\ldots,T-1$, extract $e_t$ from all observed configurations and outcomes. Rank the top/bottom three and retain the latest five; compute coverage, stagnation, and local contrasts.
  \item Ask the reflector for evidence-linked add/confirm/revise/weaken/retire operations and a proposed regime using $e_t$ and $S_t^-$. Apply validation and semantic deduplication, execute operations, apply eligible staleness decay, and govern active capacity to obtain $S_t$. Resolve the regime with the deterministic gate.
  \item Construct $N$ legal, unseen candidates from global diversity, state guidance, local exploitation, and recovery. Reserve up to two slots when $N=10$ for eligible matched interventions. Protect these tests from the shared LLM candidate refiner; validate and replenish the ordinary candidates.
  \item Query the shared LLAMBO surrogate on the current history and pool; compute EI. Select its maximizer unless a valid test-policy slot is available. An override requires an eligible test, support score at least $0.55$, and the budget/interval checks below.
  \item Evaluate the selected configuration with the true oracle and append its outcome to obtain $\mathcal D_{t+1}$. If its lineage passes direct-test eligibility, compare its outcome with the identified observed control and update only the parent hypothesis. Record global-incumbent progress separately and carry the resulting state as $S_{t+1}^-$.
  \item Store the selected configuration, true outcome, candidate lineage, state transition, and call metadata; continue until $T$ evaluations are completed.
\end{enumerate}
The matched baseline uses the same typed sampling/refinement substrate and surrogate but omits steps involving state, direct tests, and test-policy overrides. Optional domain-prior cards are disabled in both methods; task descriptions and pretrained LLM knowledge remain available.

\paragraph{Local evidence and control matching.}
For numerical variables, differences are normalized by the specified range; unordered categories use a zero/one mismatch. For a target parameter, an observational pair must have target difference at least $0.10$ and mean nuisance difference at most $0.30$. Pairs are ordered by increasing nuisance difference, decreasing target difference, and decreasing absolute outcome difference, retaining at most four per parameter and twelve overall. Median directional differences and agreement summarize support; at most five context-dependent interaction records are retained. These are observational contrasts, not identified causal effects.

A direct test has one parent hypothesis, one or two target variables, and an observed anchor. Only the primary hypothesis receives its outcome-based support update, even if other hypotheses contributed to candidate guidance. In open spaces, non-target coordinates are held fixed and targeted coordinates are changed according to the intervention. Numeric steps are $10\%$ of the range per depth (log-range for log variables); integer steps are rounded with a minimum of one, and ordered finite supports advance by their index. Depths one through six are attempted. Unordered levels change without a numerical direction. All changes are clipped/canonicalized, and unchanged or previously evaluated configurations are rejected. Interactions change the specified coordinates jointly; this is not an exhaustive factorial design. For table-backed tasks, only label-free covariate matches to observed rows can receive a direct support update, with root-mean-square and per-coordinate nuisance limits of $0.25$ and distinct oracle identities. Exact-grid candidates must be observed configurations. Approximate matches can differ in non-target variables; nearest-neighbor assignment can also change the underlying experimental row. These comparisons are not estimates of an isolated biological causal effect.

\paragraph{Ordinary candidate policies.}
Scientific-task global candidates use independent uniform sampling of numerical bounds or legal finite levels; HPO uses the stated linear/log/logit sampling scales (Appendix~\ref{sec:hpo-transfer}). Local exploitation chooses a top-evidence configuration uniformly, then samples each coordinate locally with probability $0.8$ (globally otherwise). Continuous local sampling is uniform within $12\%$ of the full range around the center, clipped to bounds; integers use the corresponding rounded radius, ordered supports use the center and adjacent levels, and unordered categories retain the center with probability $0.8$ or sample a legal level. State-guided candidates prioritize testable anchor interventions using adaptive-priority sampling, with a small nuisance perturbation for ordinary guided proposals; these are not automatically eligible for support updates. Recovery samples up to 128 global proposals to find normalized distance at least $0.25$ from repeated low-gain centers, then falls back to global sampling. Challenger proposals add evidence-linked alternative parameter/direction/context hypotheses; they do not replace every recovery draw.

Ratios are normalized and integer quotas use largest-remainder allocation. Unusable guided quotas return to global search or recovery rather than generating ungrounded tests.
\paragraph{Support score and lifecycle.}
The LLM-proposed initial support score is a validated heuristic, not a calibrated posterior probability or a test-success frequency. For new falsifiable cards, uncertainty caps it at $0.60/0.72/0.85$ for high/medium/low uncertainty. Let $v_j$ be $1$, $0.75$, or $0.5$ for low, medium, or high uncertainty, $M_{j,x}\in[0,1]$ the recorded match quality, $a_{j,x}=0.75$ for a transferred anchor and $1$ otherwise, and $n_{j,\mathrm{ctrl}}$ its previous uses. The effective evidence weight is
\begin{equation}
w_{j,x}=\operatorname{clip}\left(
\frac{v_j M_{j,x}a_{j,x}}{1+n_{j,\mathrm{ctrl}}},0,1\right).
\end{equation}
Exact open-space interventions have $M_{j,x}=1$; table-backed match quality is $\exp[-(d_{\mathrm{rms}}+d_{\mathrm{max}})]$. Approximate matches receive discounted weight. The threshold $\epsilon_{j,t}$ is the larger of the extracted effect threshold ($0.02$ times the observed score span, floor $10^{-12}$) and $\max(10^{-10},10^{-8}\max(|y^{\mathrm{ctrl}}_{j,t}|,1))$. This empirical tolerance is not a noise-calibrated significance threshold. Support changes by $+0.12w$, $0$, or $-0.08w$. Staleness applies when rounds since verification exceed the card's window (two for cards created at the early stage, three subsequently), and subtracts $0.08$ once per last-verified epoch. Active and weakened bands use $0.55$ and $0.22$; low-support cards can be archived without being retired. Failure-based retirement requires at least two consecutive failures and cumulative consecutive failure weight at least two. Historical controls are reused under a fixed response oracle; this does not control temporal drift or measurement noise in prospective experiments.

\paragraph{Adaptive priority and override accounting.}
Eligible test candidates are ordered lexicographically by adaptive priority, EI, support score times uncertainty weight and match quality, challenger flag, negative target distance, and negative candidate index. Adaptive priority is
\begin{equation}
\begin{aligned}
p_j={}&(0.5+0.5q_j)U_j B_j R_j\\
&{}\times (1+n_j)^{-1/2}(1+n_{\mathcal P_j})^{-1/2}\\
&{}\times(1+n_{\mathrm{anchor}})^{-1/2}\\
&{}\times
\frac{1}{|\mathcal P_j|}
\sum_{p\in\mathcal P_j}(1+n_p)^{-1/2}.
\end{aligned}
\end{equation}
Here $n_j$, $n_{\mathcal P_j}$, $n_{\mathrm{anchor}}$, and $n_p$ are direct-test counts for the hypothesis, parameter set, anchor, and individual parameter. $U_j$ is $0.65/0.85/1$ for low/medium/high uncertainty, $B_j=1.15$ for challengers and $1$ otherwise, and $R_j=\sqrt{(d+1)/(d+2)}$ for $d$ rounds since the more recent of the hypothesis and parameter-set tests ($1$ if untested). Saturated/ineligible contexts receive zero priority. This is a deterministic heuristic, not a learned acquisition function. The uncertainty factor in this priority differs deliberately from the conservative evidence weight.

Positive budget fractions define caps $\max(1,\lfloor0.20T\rfloor)$ and $\max(1,\lfloor0.10T\rfloor)$ for base and additional challenger slots. For $T=20$, these are four and two. Base usage counts trials receiving direct support updates, including tests selected by EI; extra slots require stagnation and exhausted base usage. Overrides have a minimum interval of three rounds. EI-selected tests are not excluded when the override budget is exhausted. The direct candidate quota is $\max(1,\operatorname{round}(0.20N))$, preserving at least one ordinary candidate when $N>1$.

\begin{table*}[t]
\centering
\small
\begin{tabular}{p{0.30\textwidth}p{0.30\textwidth}p{0.30\textwidth}}
\toprule
Setting & Value & Scope/provenance\\
\midrule
Initialization/search evaluations & Primary/HPO $10/20$; Summit $8/20$ & Six paired seeds, batch size one\\
Candidate pool; templates; generations & $10$; $2$; $10$ & Shared main-run configuration\\
Reflector output/temperature/retry & $1500$ tokens; $0$; $1$ & FGSS extra call; retries logged\\
Shared refiner output/temperature & $1000$ tokens; $0$ & Top five plus recent five observations\\
Evidence top/bottom/recent & $3/3/5$ & Outcome-ranked top/bottom\\
Stagnation windows and thresholds & $3/5$; $0.01/0.02$ & Short/long progress heuristics\\
Numeric bins; near-duplicate tolerance & $5$; $0.05$ & Normalized configuration coordinates\\
Contrast target/nuisance thresholds & $0.10/0.30$ & Observational pair selection\\
Pair limits; interactions & $4$ per parameter; $12$ total; $5$ & Extractor defaults\\
Contrast effect threshold & $0.02$ times observed score span & Floor $10^{-12}$\\
Active/per-parameter/type capacity & $4/1$; explore/exploit $2/2$ & Active guidance only\\
Support bands; stale penalty & $0.55/0.22$; $0.08$ & Not posterior probabilities\\
Success/failure increments & $+0.12/-0.08$ times evidence weight & Neutral increment zero\\
Retirement failure count/weight & $2/2$ & Both conditions required\\
Test pool/base/challenger fractions & $0.20/0.20/0.10$ & Pool quota vs override caps\\
Test interval/support score & $3$ rounds; $0.55$ & Additional eligibility checks\\
Direct nuisance bounds & $0.25/0.25$ & Aggregate and each coordinate\\
Optional domain prior & Disabled & No retrieved/static prior card\\
\bottomrule
\end{tabular}
\caption{Implementation settings inherited from development defaults and used in the recorded main runs, without a hyperparameter sweep.}
\label{tab:hyperparameters}
\end{table*}

\begin{table}[t]
\centering
\small
\begin{tabular}{lrrrr}
\toprule
Regime & Global & Guided & Local & Recovery\\
\midrule
Explore & .55 & .25 & .10 & .10\\
Mixed & .40 & .30 & .20 & .10\\
Exploit & .25 & .30 & .35 & .10\\
Recover & .45 & .15 & .15 & .25\\
\bottomrule
\end{tabular}
\caption{Nominal candidate-allocation presets, not fixed realized ratios. The gate preserves a global floor of $.30$ ($.55$ before reliable support, $.45$ under weak support); plateau recovery enforces global/recovery floors $.45/.25$ and guided cap $.20$.}
\end{table}
\paragraph{Threshold roles and configuration sources.}
The normalized-distance, support-score, and budget constants in Table~\ref{tab:hyperparameters} are shared development settings. Table~\ref{tab:reliability-cutoffs} gives fixed absolute response cutoffs $\tau$ on each task's supplied scale. These are FGSS algorithm settings, rather than biological success criteria adopted from the source studies \citep{kanda2022robotic,cosenza2022multi}. iPSC uses a CLI override; C2C12 uses its task-runner default. All six primary FGSS runs per task record the same cutoff. The effect tolerance instead adapts to the observed score span.

\begin{table}[t]
\centering
\small
\setlength{\tabcolsep}{3pt}
\begin{tabular}{lrl}
\toprule
Task & Cutoff & Configuration source\\
\midrule
iPSC-RPE & $0.12$ & Recorded CLI override\\
C2C12 & $0.05$ & Task-runner default\\
SnAr & $0.10$ & Summit task preset\\
Reizman 1--4 & $50$ & Summit task preset\\
HPLC & $1000$ & Task-runner default\\
HPO & $0$ & Task-bridge setting\\
\bottomrule
\end{tabular}
\caption{Absolute regime-gate cutoffs and their configuration sources. iPSC uses the reported override rather than the generic runner default of $0.01$.}
\label{tab:reliability-cutoffs}
\end{table}

Strong reliability requires either improvement over the initial best \emph{or} an absolute-cutoff hit, together with at least five search observations and three observations near the current best. The near-best margin is $\max(0.10\,\mathrm{span}(y),10^{-8})$ over observed outcomes. Cutoff hits mean $b_t\geq\tau$ for maximization or $b_t\leq\tau$ for minimization. Positive MSE therefore reaches the strong regime through improvement and supporting observations rather than HPO's zero cutoff.

For maximization, the same cutoff also replaces the refiner's hypothesis-guidance text with a broad-exploration instruction when both the best outcome and observed span are below $\tau$, with at most two distinct positive outcomes after six-decimal rounding. State records and the regime-controlled candidate policy remain enabled.

\subsection{Realized Test Policy and Matching}
\label{sec:test-policy-audit}
An actual override changes the selected candidate index from the recorded EI choice, rather than merely activating arbitration. Table~\ref{tab:realized-tests} audits all six full-FGSS primary runs. iPSC has 18 overrides out of 120 search evaluations ($15\%$), plus 19 EI-selected tests receiving support updates. C2C12 has neither overrides nor direct support updates: its observed gain cannot be attributed to the matched-test path. Its structured reflection and ordinary state-guided search remain enabled.

\begin{table}[t]
\centering
\small
\begin{tabular}{lrrrrrr}
\toprule
Seed & 0 & 1 & 2 & 3 & 4 & 5\\
\midrule
iPSC overrides & 2 & 4 & 2 & 4 & 3 & 3\\
iPSC EI tests & 6 & 0 & 6 & 2 & 1 & 4\\
C2C12 overrides & 0 & 0 & 0 & 0 & 0 & 0\\
C2C12 EI tests & 0 & 0 & 0 & 0 & 0 & 0\\
\bottomrule
\end{tabular}
\caption{Actual selection changes and EI-selected matched tests; each run has twenty search evaluations. EI tests receive a recorded support update without changing EI selection.}
\label{tab:realized-tests}
\end{table}

All 37 iPSC updates use approximate covariate matches and one primary hypothesis. Their match quality has median $0.748$, interquartile range $[0.729,0.784]$, and range $[0.664,0.918]$. Normalized nuisance RMS distance has median $0.093$ and maximum $0.159$; maximum-coordinate distance has median $0.188$ and maximum $0.250$.

Generation logs contain 482 iPSC validation-check events, including 155 explicitly rejected events ($32.2\%$: 152 same-oracle and three already-observed configurations). C2C12 records 476 events, including 329 same-oracle rejections ($69.1\%$). The denominators include intermediate deferred checks and repeated proposal attempts; these are logged check-event rates, not rejection rates over unique tests or all internal generation attempts. Subsequent reanchoring or observed-row matching can still succeed.

FGSS w/o Tests disables test generation, override selection, and outcome support updates together. A separate EI-only selection ablation and post-hoc fixed-oracle cost audit appear in Appendix~\ref{sec:test-selection-effects}.

\subsection{Search Spaces and Initialization}
\begin{table*}[t]
\centering
\small
\begin{tabular}{lp{0.20\textwidth}lp{0.39\textwidth}}
\toprule
Task & Variable & Type & Admissible domain\\
\midrule
SnAr & Residence time & Continuous & $[0.5,2.0]$ min\\
& Pyrrolidine equivalents & Continuous & $[1,5]$ equivalents\\
& DFNB concentration & Continuous & $[0.1,0.5]$ mol/L\\
& Temperature & Continuous & $[30,120]$ degrees Celsius\\
\midrule
Reizman 1--4 & Residence time & Continuous & $[60,600]$ s\\
& Temperature & Continuous & $[30,110]$ degrees Celsius\\
& Catalyst loading & Continuous & $[0.5,2.5]$ mol\%\\
& Catalyst identity & Categorical & P1-L1, P2-L1, P1-L2, P1-L3, P1-L4, P1-L5, P1-L6, P1-L7\\
\midrule
iPSC-RPE & PC & Ordered finite support & 54 finite levels, $12.47$--$505.62$\\
& PP; DS; KP; 3P & Integer & $[1,6]$; $[10,100]$; $[1,19]$; $[3,19]$\\
& DP; DL & Ordered finite support & $\{5,8,11,14,17,20,23\}$; $\{0,1\}$\\
\midrule
C2C12 & Media components 1--14 & Continuous & Each $[0,1]$ in adapter-normalized coordinates\\
& Information source & Finite support & $\{0,1,2,3\}$; retained from the historical offline protocol\\
\midrule
HPLC & Sample/additional/tubing volume & Continuous & $[0,.08]$; $[0,.06]$; $[.1,.9]$ ml\\
& Sample flow; push speed; wait time & Continuous & $[.5,2.5]$ ml/min; $[80,150]$ Hz; $[.5,10]$ s\\
\bottomrule
\end{tabular}
\caption{Domains used by the actual adapters. Summit bounds and catalyst levels follow benchmark definitions \citep{felton2021summit}; HPLC bounds follow the Olympus dataset configuration \citep{hase2021olympus}. iPSC concentration levels follow the original task-code definition; the adapter domains are not assumed exhaustive physical domains. C2C12 coordinates follow its adapter preprocessing.}
\label{tab:search-spaces}
\end{table*}
Summit initializes continuous coordinates with seeded Latin hypercube sampling and balances catalyst categories; initial points are identical within each paired seed. The primary iPSC-RPE and C2C12 tasks use outcome-conditioned low-score initialization and nearest-neighbor response lookup for retrospective evaluation. Exact-grid iPSC restricts candidate configurations to observed rows. HPLC initializes uniformly from observed configurations without labels and uses the reported offline lookup. C2C12's source indicator denotes measurement provenance, so the optimizer queries configuration--source pairs.

\paragraph{C2C12 table provenance and response scale.}
\label{sec:c2c12-provenance}
The processed table copies all 209 rows of the upstream DBO archive: the first fourteen columns of \texttt{BO\_data.txt} provide media coordinates already on $[0,1]$, and its fifteenth column supplies \texttt{source}. The two columns of \texttt{BO\_outputs.txt} are retained verbatim as \texttt{score} and \texttt{score\_var}. An all-row numerical audit confirms this mapping exactly. The source study measures C2C12 growth through cell counting and correlated assay responses; we report the saved response-score scale rather than reconstructing absolute cell counts or an assay-specific inverse transformation. Maximization seeks a larger archived response at a fixed evaluation budget. Media costs, source-specific evaluation costs, and source-bias calibration are outside this objective.

All rows define a fixed response lookup, with no C2C12 train/test split or learned emulator fitted in this study. Beyond the disclosed low-score initialization, surrogate and reflector inputs contain queried outcomes only. Matching unobserved configurations and interventions consults covariates, not their response labels. The oracle returns a saved scalar deterministically; \texttt{score\_var} is retained as metadata rather than used to sample additional observation noise.

\subsection{Reflector Prompt and State Schema}
\label{sec:reflector-state}
The reflector receives current evidence and the previous state, and returns an assessment, a proposed search regime, and evidence-linked operations. The following is an abbreviated \emph{output contract}, not a claimed model response:
{\scriptsize
\begin{verbatim}
{
  "global_assessment": {
    "stage": "early|mid|late",
    "search_mode_suggestion": "balanced",
    "uncertainty_note": "..."
  },
  "operations": [
    {
      "op": "add",
      "type": "exploit",
      "parameter": "<legal parameter>",
      "parameters": ["<legal parameter>"],
      "hypothesis_kind": "main_effect",
      "anchor_observation_id": "obs_...",
      "intervention": {"changes": {...},
        "controlled_parameters": [...]},
      "expected_outcome": {
        "objective": "improve",
        "direction": "increase"},
      "evidence_pair_ids": ["pair_..."],
      "evidence_refs": ["contrastive_evidence"],
      "confidence": 0.63
    }
  ],
  "search_phase": {
    "mode": "explore|mixed|exploit|recover",
    "target_parameters": [...]
  }
}
\end{verbatim}
}
Existing-card operations use \texttt{target\_id}; the early prompt permits only add operations, while later prompts also permit confirm, revise, weaken, and retire. The validator requires evidence references and executable ranges. Invalid responses can trigger a rule-based fallback; deterministic repair and challenger operations can supplement valid responses, and these paths are recorded. A falsifiable test specification cannot be silently changed in place; a different intervention requires a new evidence-linked card. The LLM does not supply measured test counts or invent tested candidate IDs: these are populated by the deterministic evaluation/support-update path. Authentic schema and log keys retain \texttt{confidence} for the heuristic support score and \texttt{credit} for comparative evidence updates; neither denotes a calibrated probability or an identified causal effect.

\paragraph{Recorded state example.}
The following fields are selected from the pre-selection state in Reizman Case 3, seed 0, search round 8 (zero-based). Values and IDs are retained from the saved log.
{\scriptsize
\begin{verbatim}
{
  "id": "llm_h_6_14",
  "parameters": ["temperature"],
  "anchor_observation_id": "obs_000000",
  "evidence_pair_ids": [
    "pair_temperature_0000_0011"],
  "confidence": 0.6399999999999999,
  "validation_status": "partially_supported",
  "tested_count": 1,
  "success_count": 1,
  "last_test_credit_weight": 0.5,
  "last_test_improvement": 5.174800627842362,
  "global_improvement_count": 0,
  "hypothesis_value_status":
    "locally_supported"
}
\end{verbatim}
}
The same round proposed exploitation but resolved to mixed under weak reliability. The identified card and its recorded test outcomes illustrate the state objects that guide search and retain comparative support; local support remains distinct from global optimization progress.

\subsection{API Settings and Recorded Usage}
The requested backend is \texttt{gpt-4o-mini}; the reflector submits its assembled instruction and evidence as a user message, at temperature zero with a 1500-token output cap and one configured retry. It does not explicitly set \texttt{top\_p} or an API sampling seed. Optimization seeds and the feedback-permutation seed are separate from API randomness. Temperature zero does not guarantee identical cloud responses; deterministic curation and validation refer to fixed recorded inputs, not the entire optimizer.

Each primary task has 120 final reflector records marked as successful LLM parses, with no logged fallback. Summed recorded reflector usage is $2{,}079{,}239$ prompt and $62{,}333$ completion tokens for iPSC, and $2{,}403{,}748$ prompt and $69{,}019$ completion tokens for C2C12. These are reflector-only usage totals, not total optimizer compute or billing. Final call records do not supply a complete count of transient failed requests or retries.

\begin{table}[t]
\centering
\footnotesize
\setlength{\tabcolsep}{3pt}
\begin{tabular}{llrrr}
\toprule
Task & Seed & \shortstack{Prompt\\tokens} & \shortstack{Completion\\tokens} & \shortstack{Elapsed\\(s)}\\
\midrule
iPSC & 0 & 347576 & 10226 & 177.5\\
& 1 & 332143 & 10664 & 144.0\\
& 2 & 368290 & 10319 & 144.7\\
& 3 & 346650 & 10845 & 145.7\\
& 4 & 346886 & 10709 & 127.8\\
& 5 & 337694 & 9570 & 113.4\\
\midrule
C2C12 & 0 & 412771 & 11030 & 175.7\\
& 1 & 365428 & 12230 & 178.8\\
& 2 & 370151 & 10958 & 156.0\\
& 3 & 361268 & 10461 & 161.2\\
& 4 & 457476 & 11976 & 168.6\\
& 5 & 436654 & 12364 & 174.1\\
\bottomrule
\end{tabular}
\caption{Recorded full-FGSS reflector usage per primary-task run, with twenty final call records each. Elapsed is the sum of saved reflector-call latencies, not total optimizer wall-clock time.}
\label{tab:reflector-usage}
\end{table}

Historical records do not identify exact served model snapshots or per-request timestamps; temperature-zero calls do not guarantee bitwise cloud-API replay.

\subsection{Non-LLM Baseline Implementations and Trajectories}
\label{sec:nonllm-details}
Random Search samples numerical bounds or finite legal levels uniformly. HEBO uses version 0.3.5, with the imported initialization count determining its startup size. In the primary tasks its GP fitter uses Adam instead of the default pSGLD after a numerical failure in C2C12 smoke; other model settings are retained. Ordered finite levels are encoded by index and unordered labels as categories. This fitting change is not applied to the Summit runs, which retain the package defaults. Cross-domain differences therefore cannot be attributed solely to task domain.

GP-EI uses BoTorch 0.8.5 with standardized outcomes and range-normalized numerical coordinates. It uses a categorical covariance component for iPSC pipetting length, C2C12 source, and Reizman catalyst identity. Acquisition optimization uses ten restarts and 256 raw samples; continuous optimization has a maximum of 200 iterations. Fully discrete iPSC uses legal discrete local search, C2C12 enumerates source labels, and Summit uses continuous or mixed optimization as appropriate. No wrapper-level random fallback replaces failed fitting. Repeated evaluations consume budget, and all methods reuse the actual paired initialization.

\begin{figure*}[t]
  \centering
  \includegraphics[width=\textwidth]{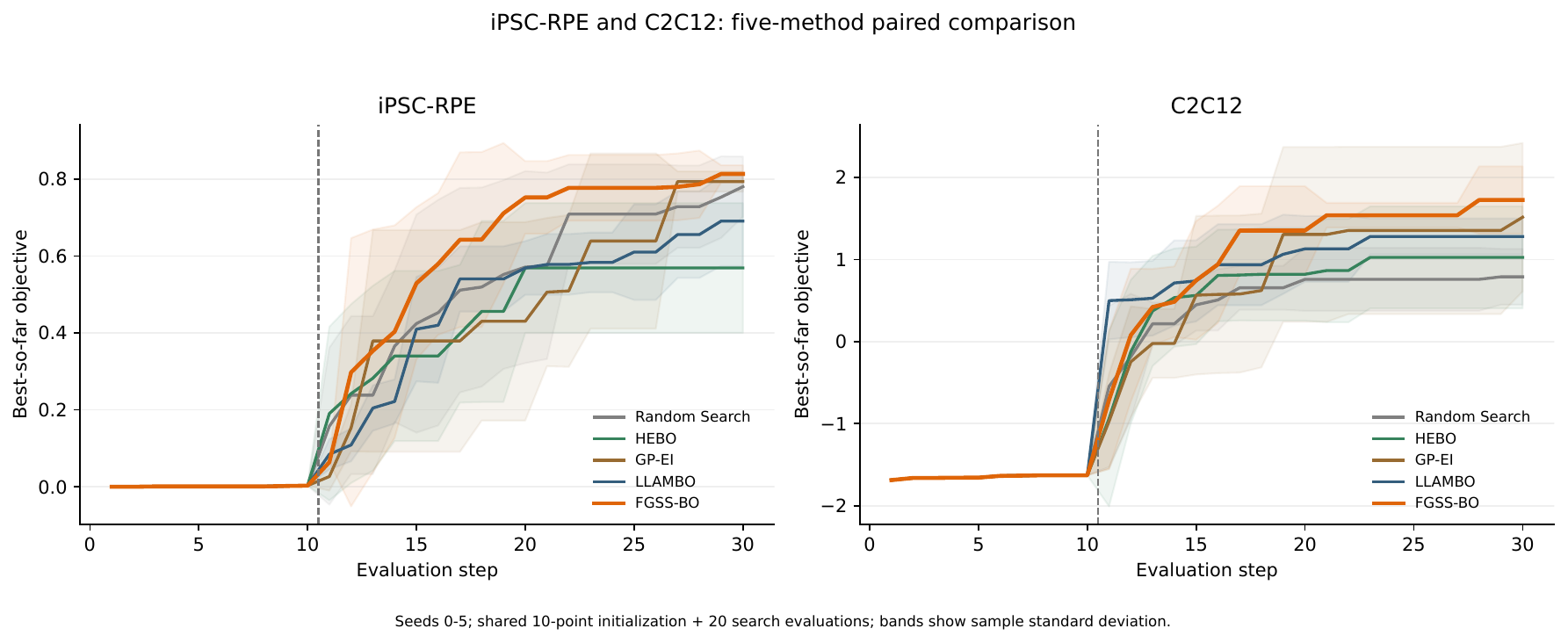}
  \caption{Five-method primary-task trajectories, six paired runs with ten shared low-score initialization observations and twenty search evaluations. Lines are means and bands are sample standard deviations. All methods use the same nearest-neighbor response oracle; bands are not confidence intervals. LLAMBO uses the shared feasibility-preserving interface.}
  \label{fig:biological-fiveway}
\end{figure*}

\subsection{Paired Primary-Task Differences}
\label{sec:paired-differences}
Table~\ref{tab:paired-differences} reports FGSS-BO minus the matched LLM and GP-EI references. Intervals use 10,000 paired bootstrap resamples of the six run-level differences (seed 42) and are exploratory, without multiplicity correction.

\begin{table*}[t]
\centering
\small
\begin{tabular}{lllrr}
\toprule
Task & Reference & Metric & Mean difference & Bootstrap 95\% interval\\
\midrule
iPSC-RPE & GP-EI & Final best & $0.020$ & $[-0.009,0.050]$\\
iPSC-RPE & LLAMBO & Final best & $0.123$ & $[0.046,0.205]$\\
iPSC-RPE & GP-EI & Search AUC & $0.135$ & $[0.014,0.254]$\\
iPSC-RPE & LLAMBO & Search AUC & $0.147$ & $[0.054,0.252]$\\
C2C12 & GP-EI & Final best & $0.210$ & $[-0.271,0.691]$\\
C2C12 & LLAMBO & Final best & $0.445$ & $[0.110,0.959]$\\
C2C12 & GP-EI & Search AUC & $0.299$ & $[-0.248,0.779]$\\
C2C12 & LLAMBO & Search AUC & $0.141$ & $[-0.134,0.419]$\\
\bottomrule
\end{tabular}
\caption{Paired primary-task differences, FGSS-BO minus the reference; positive favors FGSS-BO. LLAMBO uses the shared feasibility-preserving interface.}
\label{tab:paired-differences}
\end{table*}

\subsection{Cross-Domain Summit Evaluation}
\label{sec:summit-cross-domain}
We evaluate five Summit tasks. SnAr is a continuous kinetic-flow simulator; Reizman Cases 1--4 are continuous/categorical reaction emulators with catalyst choices. Each method shares eight initial observations from the LLM comparison, followed by twenty search evaluations, over six paired seeds. Table~\ref{tab:summit-cross-domain} compares all five methods. HEBO and GP-EI have higher sample mean final best and search AUC than FGSS-BO and LLAMBO on every task. FGSS-BO exceeds its matched LLM reference on four final-best means and three AUC means, but underperforms it on Case 4.

\paragraph{Objectives and oracle implementation.}
All five objectives are maximized. SnAr uses $\mathrm{STY}/10^4-E_{\mathrm{factor}}/100$, the sign reversal of the benchmark tutorial's minimization objective. Space-time yield (STY) is measured in $\mathrm{kg}\,\mathrm{m}^{-3}\,\mathrm{h}^{-1}$; E-factor is the waste-to-product mass ratio. The scalar score combines throughput and waste, rather than reporting reaction yield. Reizman Cases 1--4 maximize predicted yield in percent, with responses clipped to $[0,100]$. They share three continuous controls and an eight-level catalyst variable; names, units, and bounds appear in Table~\ref{tab:search-spaces}.

The downloaded Summit source distribution declares version 0.8.9. Our adapter implements the zero-noise SnAr kinetics with SciPy's \texttt{solve\_ivp} and reads the official JSON preprocessing statistics and five fixed PyTorch predictors for each \texttt{reizman\_suzuki\_case\_1}--\texttt{4} model directly. No emulator is retrained or additional observation noise sampled. Run seeds control the paired Latin hypercube initialization, category balancing, and candidate sampling; the oracle remains fixed, and cloud LLM randomness is separate. An offline replay of all 1680 saved responses from the 60 LLM runs agrees with these assets within $1.5\times10^{-14}$. Metrics are reported per task without treating declared output bounds as verified global optima.

\begin{table*}[t]
  \centering
  \footnotesize
  \setlength{\tabcolsep}{3pt}
  \begin{tabular}{llccccc}
    \toprule
    \textbf{Task} & \textbf{Metric} & \shortstack{\textbf{Random}\\\textbf{Search}} & \textbf{HEBO} & \textbf{GP-EI} & \textbf{LLAMBO} & \textbf{FGSS-BO} \\
    \midrule
    SnAr & Final best & $0.593\pm0.074$ & $1.051\pm0.005$ & $1.058\pm0.001$ & $0.641\pm0.107$ & $0.680\pm0.057$ \\
    SnAr & Search AUC & $0.572\pm0.078$ & $0.967\pm0.048$ & $0.990\pm0.049$ & $0.579\pm0.101$ & $0.573\pm0.123$ \\
    \midrule
    Reizman Case 1 & Final best & $67.505\pm10.443$ & $86.480\pm6.887$ & $91.424\pm0.019$ & $60.071\pm15.792$ & $69.751\pm14.764$ \\
    Reizman Case 1 & Search AUC & $64.033\pm9.672$ & $74.673\pm9.317$ & $83.315\pm5.291$ & $55.087\pm13.751$ & $65.738\pm15.484$ \\
    \midrule
    Reizman Case 2 & Final best & $21.602\pm6.882$ & $37.151\pm6.817$ & $40.085\pm0.000$ & $20.910\pm7.124$ & $22.286\pm9.016$ \\
    Reizman Case 2 & Search AUC & $19.916\pm6.719$ & $32.166\pm6.307$ & $32.762\pm4.777$ & $19.164\pm6.721$ & $21.262\pm8.662$ \\
    \midrule
    Reizman Case 3 & Final best & $79.543\pm5.887$ & $89.536\pm2.604$ & $92.770\pm0.241$ & $73.964\pm8.452$ & $76.606\pm4.921$ \\
    Reizman Case 3 & Search AUC & $72.094\pm3.800$ & $85.115\pm3.114$ & $84.723\pm4.589$ & $68.152\pm3.466$ & $71.726\pm2.183$ \\
    \midrule
    Reizman Case 4 & Final best & $90.216\pm3.366$ & $95.541\pm2.615$ & $96.688\pm0.034$ & $92.000\pm4.124$ & $89.866\pm3.561$ \\
    Reizman Case 4 & Search AUC & $88.079\pm2.992$ & $93.910\pm2.093$ & $94.243\pm1.450$ & $89.939\pm2.651$ & $88.505\pm3.597$ \\
    \bottomrule
  \end{tabular}
  \caption{Cross-domain Summit results: five methods over six paired runs, with eight initialization and twenty search evaluations. Values are mean $\pm$ sample standard deviation; higher is better. LLAMBO uses the shared feasibility-preserving interface.}
  \label{tab:summit-cross-domain}
\end{table*}

\subsection{Exact-Grid iPSC-RPE Control}
\label{sec:exact-grid}
The exact-grid variant restricts candidates to observed experimental configurations and does not assign new configurations nearest-neighbor responses. LLAMBO obtains final best $0.840\pm0.026$ and search AUC $0.750\pm0.009$; FGSS-BO obtains $0.848\pm0.064$ and $0.710\pm0.096$. The final-best means are close and the AUC advantage is not preserved. This control changes the admissible candidate set as well as removing nearest-neighbor assignment, so it does not isolate oracle effects from candidate-space restriction.

\subsection{Olympus HPLC Control}
\label{sec:hplc-control}
The continuous-parameter Olympus HPLC benchmark \citep{hase2021olympus} provides a further experimental-design task. Across six paired runs, LLAMBO obtains final best $1803.720\pm521.827$ and search AUC $1677.365\pm465.783$; FGSS-BO obtains $1962.651\pm237.131$ and $1765.846\pm381.965$. FGSS-BO has higher sample means on both metrics; this comparison covers the two LLM pipelines.
\subsection{Component-Ablation Protocol and Paired Differences}
\label{sec:component-details}
All four iPSC variants share ten low-score initial observations, twenty search evaluations, the legal parameter domains, and the nearest-neighbor response oracle. The shared LLM settings are GPT-4o mini, ten candidates, two templates, and ten generations. The two reflective controls and full FGSS use a reflection output cap of 1500 tokens, temperature zero, and one configured retry. This matches call settings, not actual tokens or total compute with LLAMBO.

\paragraph{Reflective Summary.}
Each round, the same structured evidence extractor supplies current evidence alongside the previous prose summary. The LLM replaces that summary with updated guidance on promising directions, contradictions, uncertainty, coverage, and exploration. The summary conditions the shared candidate refiner, without hypothesis cards, deterministic hypothesis curation, matched-test generation, direct support updates, or test-policy overrides. Both this control and FGSS maintain cross-round memory under the shared reflection settings.

\paragraph{FGSS w/o Tests.}
Structured reflection, curation, search-regime gating, and ordinary state-guided candidates remain enabled. Matched-test generation, test-policy overrides of EI, and direct-test support updates are disabled together. This is a testing-bundle ablation, not an isolation of selection overrides alone. All twenty selections use the shared EI path. Realized full-FGSS selection changes and control-match quality appear in Appendix~\ref{sec:test-policy-audit}.

\paragraph{Reference reuse and execution audit.}
The LLAMBO and full-FGSS references reuse the primary iPSC runs. A source-compatibility audit checked the operative extractor, curator, reflector instructions, phase gate, candidate policies, surrogate, and oracle paths; subsequent additions were optional interfaces and logging. Paired initialization, objective outcomes, and direct-control scores were also checked. All runs completed the required thirty evaluations and remain in the analysis, including the run with one FGSS w/o Tests reflection (seed 4, search round 19, one-based) that used the recorded rule fallback after a connection failure.

Table~\ref{tab:component-differences} uses 50,000 paired bootstrap resamples with analysis seed 20261012. The intervals are exploratory percentile intervals without multiplicity correction. Full FGSS has a higher observed final-best mean than the summary, but no AUC advantage; removing the test bundle has mixed effects. The contrast compares complete prose-summary and structured pipelines, which differ in state representation, candidate control, and test management.

\begin{table*}[t]
\centering
\small
\begin{tabular}{llrr}
\toprule
Contrast & Metric & Mean difference & Bootstrap 95\% interval\\
\midrule
Full FGSS $-$ Reflective Summary & Final best & $0.0285$ & $[0.0050,0.0617]$\\
& Search AUC & $-0.0061$ & $[-0.0702,0.0605]$\\
Full FGSS $-$ FGSS w/o Tests & Final best & $0.0088$ & $[-0.0370,0.0730]$\\
& Search AUC & $-0.0275$ & $[-0.1264,0.0651]$\\
FGSS w/o Tests $-$ Reflective Summary & Final best & $0.0197$ & $[-0.0527,0.0910]$\\
& Search AUC & $0.0214$ & $[-0.0310,0.0822]$\\
\bottomrule
\end{tabular}
\caption{Paired iPSC component differences over six runs. Positive favors the first method.}
\label{tab:component-differences}
\end{table*}

\subsection{State-Feedback Permutation on iPSC-RPE}
\label{sec:state-feedback-permutation}
This control retains the full FGSS mechanism but permutes historical outcomes across configurations only for state construction and updates. Configurations, initialization, domains, and evaluation budgets are unchanged. LLAMBO's surrogate and the shared candidate refiner retain true observation history; the refiner also receives the perturbed state guidance. The permutation is refreshed at pre- and post-evaluation stages, with feedback seed 42. Each eligible support comparison uses the same post-evaluation view for both intervention and control. All reported performance metrics use true oracle outcomes.

\begin{figure*}[t]
\centering
\includegraphics[width=\textwidth]{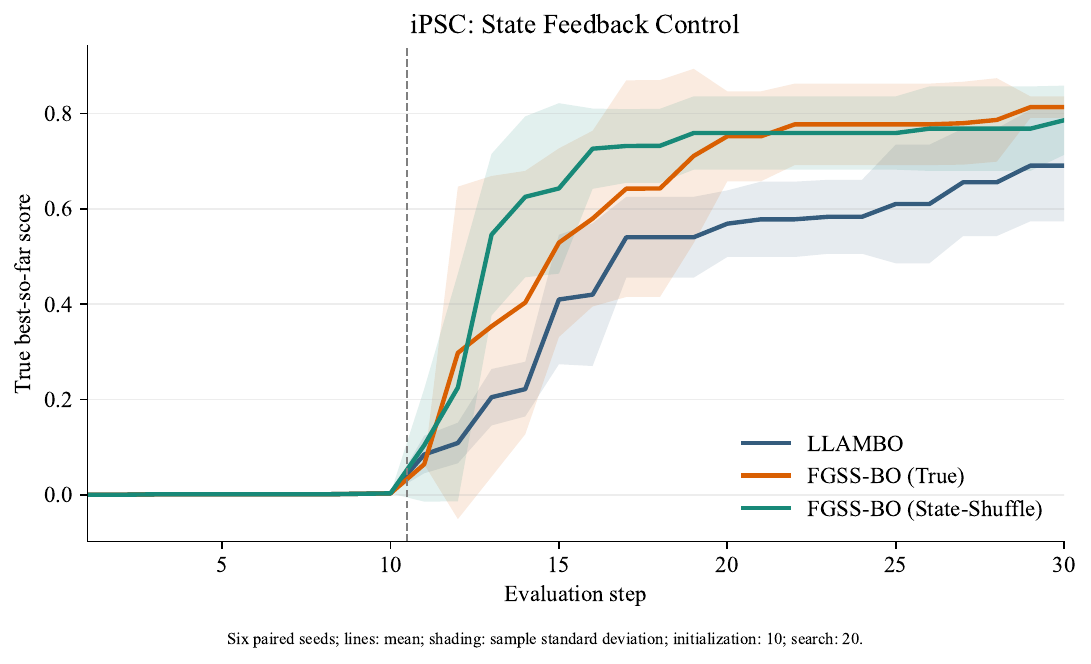}
\caption{iPSC-RPE with true or permuted state feedback, alongside the LLAMBO reference. Lines show mean best-so-far true objective over six paired runs; shading shows one sample standard deviation. The dashed boundary separates ten initialization observations from twenty search evaluations. Only FGSS-BO state feedback is permuted; the surrogate retains true observation history.}
\label{fig:state-feedback-permutation}
\end{figure*}

Figure~\ref{fig:state-feedback-permutation} includes all six paired seeds 0--5. The true-FGSS and LLAMBO references reuse the primary runs; the permutation runs were completed separately under the same experimental settings. The audit verifies paired initialization and true surrogate-history routing, and all 30 direct support updates agree with their single-view comparisons.

\begin{table}[t]
\centering
\small
\begin{tabular}{lrr}
\toprule
Condition & Final best & Search AUC\\
\midrule
True & $.8132\pm.0225$ & $.6401\pm.1175$\\
Permuted & $.7857\pm.0726$ & $.6751\pm.0654$\\
\bottomrule
\end{tabular}
\caption{iPSC-RPE state-feedback control: mean $\pm$ sample standard deviation over six paired runs.}
\label{tab:state-feedback-permutation}
\end{table}

State-feedback permutation achieves faster early progress and a higher mean search AUC, whereas aligned feedback achieves a higher mean final objective with lower across-seed variability in final outcomes (Table~\ref{tab:state-feedback-permutation}; Figure~\ref{fig:state-feedback-permutation}). This pattern is consistent with a trade-off between early progress and final-result consistency. The true-minus-permuted mean differences are $0.0275$ for final best and $-0.0349$ for AUC; their exploratory paired bootstrap 95\% intervals are $[-0.0232,0.0850]$ and $[-0.1236,0.0631]$, respectively. Both include zero. The surrogate retains true observation history in both conditions.

\subsection{Test-Selection Ablation and Immediate Opportunity Cost}
\label{sec:test-selection-effects}
\paragraph{EI-only selection control.}
FGSS No-Override retains the full reflector, curator, state-guided candidate construction, matched-test candidates, and direct support updates. Only the final selection policy changes to \texttt{ei\_only}; the test-generation quota is retained. All six seeds 0--5 complete the same ten low-score initialization observations and twenty search evaluations. They share initialization, legal domains, oracle, and LLM settings with the references in Table~\ref{tab:selection-ablation}. The audit checks all 120 selected configurations against the CSVs and confirms that each maximizes saved predicted EI. The pools contain 99 test-candidate appearances, and 22 selected tests receive a single-parent support update; all 22 control outcomes match prior observations and use distinct oracle identities. All 120 final reflector records are successful LLM calls without logged fallback.

\begin{table*}[t]
\centering
\small
\begin{tabular}{lrrrr}
\toprule
Method & Final best & Search AUC & Selection changes & Direct support updates\\
\midrule
LLAMBO & $0.6905\pm0.1168$ & $0.4936\pm0.0571$ & 0 & 0\\
Full FGSS & $0.8132\pm0.0225$ & $0.6401\pm0.1175$ & 18 & 37\\
FGSS No-Override & $0.8015\pm0.1103$ & $0.6896\pm0.0980$ & 0 & 22\\
\bottomrule
\end{tabular}
\caption{iPSC selection-policy control over six paired runs: mean $\pm$ sample standard deviation. Counts are totals over 120 search evaluations per method. Selection changes mean departures from the recorded EI choice, not tests naturally selected by EI. Full FGSS and LLAMBO reuse the audited primary references and shared feasibility-preserving interface.}
\label{tab:selection-ablation}
\end{table*}

\begin{figure*}[t]
\centering
\includegraphics[width=\textwidth]{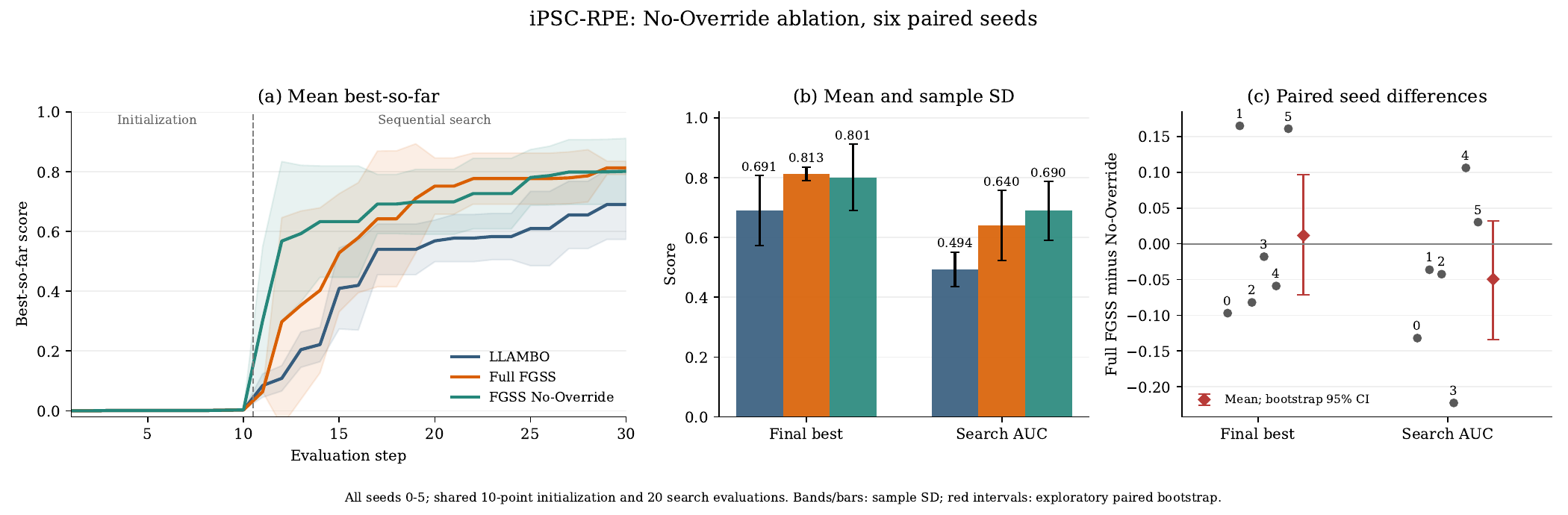}
\caption{iPSC No-Override ablation, all six paired seeds. Curves and bars show means; shading and bar error bars show sample standard deviations. The right panel reports full FGSS minus No-Override: numbered points are individual seeds, and red intervals are exploratory paired bootstrap 95\% intervals. No-Override retains hypothesis-test generation and support updates but always selects by EI. LLAMBO uses the shared feasibility-preserving interface.}
\label{fig:no-override}
\end{figure*}

Full FGSS minus No-Override has mean paired differences $+0.0117$ in final best and $-0.0495$ in search AUC. The respective exploratory bootstrap 95\% intervals are $[-0.0712,0.0965]$ and $[-0.1343,0.0324]$, from 50,000 paired-seed resamples with analysis seed 20261012. Both intervals include zero. Full FGSS has lower observed final-best dispersion, while No-Override has higher mean search AUC (Figure~\ref{fig:no-override}). No-Override's higher sample means than LLAMBO show that gains occur without forced test selection in these runs. Collection periods and cloud API randomness remain uncontrolled.

\paragraph{Same-history, same-pool opportunity cost.}
For each of full FGSS's 18 actual selection changes, we query the original fixed oracle after the fact on the recorded EI choice and selected test, without changing history or generating a new pool. Let $v_t^{\mathrm{EI}}=\widehat f_{\mathrm{NN}}(x_{t+1}^{\mathrm{EI}})$ and $v_t^{\mathrm{test}}=\widehat f_{\mathrm{NN}}(x_{t+1}^{\mathrm{test}})$, with $b_t$ the incumbent before that evaluation. Define
\begin{equation}
\begin{aligned}
g_t^{\mathrm{resp}} &= v_t^{\mathrm{EI}}-v_t^{\mathrm{test}},\\
g_t^{\mathrm{best}} &= \max(b_t,v_t^{\mathrm{EI}})\\
&\quad-\max(b_t,v_t^{\mathrm{test}}).
\end{aligned}
\end{equation}
Positive values favor the EI alternative; negative values favor the test. Replaying all 180 recorded responses (initialization included) reproduces the historical scores, and all selected-test scores and saved EI maxima agree with their logs. No LLM calls or new laboratory evaluations are performed.

\begin{figure*}[t]
\centering
\includegraphics[width=\textwidth]{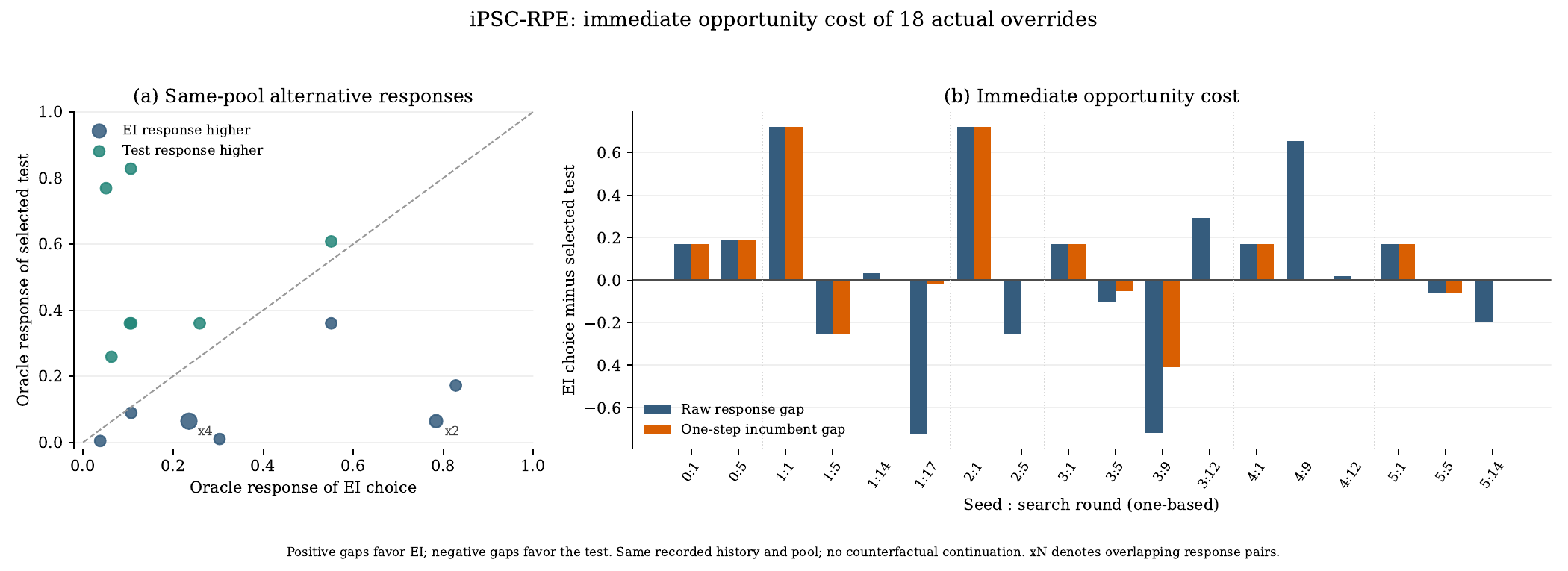}
\caption{Immediate opportunity costs of all 18 full-FGSS selection changes under the fixed iPSC oracle. The left panel compares the alternative response scores; the right reports EI minus test response and one-step incumbent gaps, indexed by seed and one-based search round. Positive gaps favor EI. Each comparison uses the same recorded history and candidate pool.}
\label{fig:immediate-cost}
\end{figure*}

Across all 18 events, mean response gap is $0.0563$: EI has higher response in 11 events and the test in seven. Mean one-step incumbent gap is $0.0854$: EI is higher in eight events, the test in five, with five ties (Figure~\ref{fig:immediate-cost}). These are event-weighted means; whole-seed cluster bootstrap, rather than independent-event resampling, gives exploratory 95\% intervals $[-0.0433,0.1932]$ and $[0.0001,0.1964]$, respectively. Intervals use all events and six seed clusters, without multiplicity adjustment.

As a post-hoc descriptive partition, the six first-search-round changes all favor EI immediately, with mean gap $0.3540$ for both measures. The remaining twelve events have mean response/incumbent gaps $-0.0926/-0.0489$. All 18 events remain included in the aggregates. This audit compares one-step outcomes under a fixed oracle; it does not estimate downstream effects.

\subsection{Freeform and Feasible Candidate Interfaces}
\label{sec:freeform-interface}
\begin{figure*}[t]
\centering
\includegraphics[width=\textwidth]{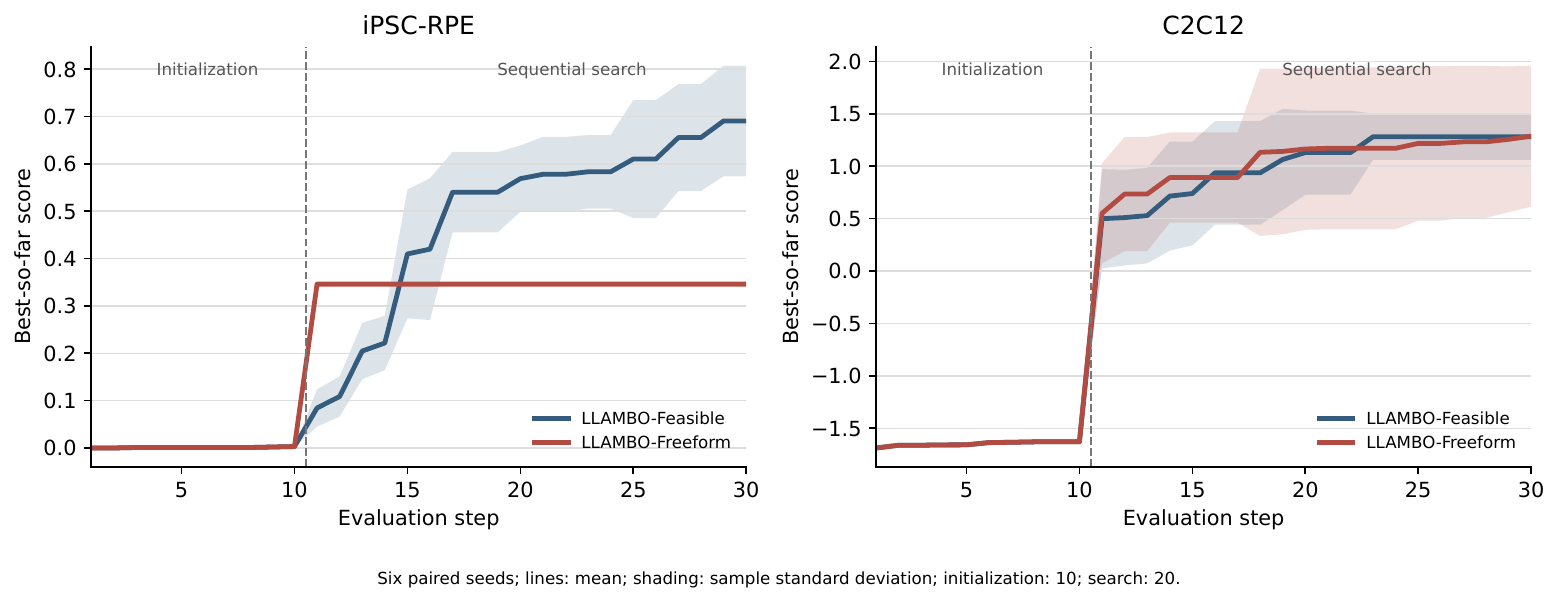}
\caption{Candidate-interface controls on both primary tasks over all six paired seeds. Lines are means; shading is sample standard deviation. The dashed line separates ten shared low-score initial observations from twenty search evaluations. iPSC Freeform uses recovery fallback in all 120 search rounds, producing six identical evaluation trajectories; C2C12 has no recovery fallback. Both methods omit FGSS state.}
\label{fig:freeform-feasible}
\end{figure*}

Figure~\ref{fig:freeform-feasible} compares LLAMBO-Freeform with the existing LLAMBO-Feasible references on iPSC-RPE and C2C12, using all seeds 0--5. All 24 runs complete ten initialization and twenty search evaluations with batch size one. Paired initial configurations and outcomes, parameter domains, and task-specific nearest-neighbor protocols agree. The Freeform runs use GPT-4o mini with ten candidates, two templates, and ten generations; neither method uses FGSS state or a reflector.

On iPSC, Freeform obtains $0.346$ in both final best and search AUC, versus Feasible's $0.691\pm0.117$ and $0.494\pm0.057$. All 120 Freeform search rounds exhaust the free-form proposal path and select recovery-fallback candidates; the six evaluation CSVs are identical. Its zero dispersion reflects coincident fallback trajectories, not reliable LLM optimization. On C2C12, no search round invokes recovery fallback: Freeform obtains $1.287\pm0.673$ and $1.058\pm0.583$, versus Feasible's $1.281\pm0.220$ and $1.025\pm0.264$. Their sample means are close, with larger Freeform dispersion. Search AUC excludes initialization.

These controls compare complete candidate interfaces. Feasible includes candidate refinement; Freeform retains validity checks, deduplication, retries, and recovery. The latter is an implementation-level free-form variant, not a strict reproduction of original LLAMBO. The new Freeform runs record their settings, whereas older Feasible references lack complete per-run CLI provenance; collection periods and cloud API responses are not controlled by initialization pairing. Feasibility-preserving construction addresses iPSC's severe proposal rejection, while C2C12's viable Freeform path yields sample means close to Feasible. Both main LLM methods use the same feasible interface across tasks; Freeform is reported separately as a candidate-interface control.

\begingroup
\setlength{\textfloatsep}{12pt plus 2pt minus 2pt}
\setlength{\dbltextfloatsep}{12pt plus 2pt minus 2pt}
\subsection{Direct-Training Bayesmark HPO Transfer}
\label{sec:hpo-transfer}
The suite was fixed before evaluation: breast cancer, handwritten digits, and diabetes datasets, each paired with RandomForest and SVM. All six tasks and both methods are retained. There are 72 completed runs and 2160 configuration evaluations, with six paired seeds, ten initialization points, twenty search evaluations, and batch size one. Table~\ref{tab:hpo-transfer} reports CV-based optimization metrics. The suite compares the two complete LLM pipelines.

\begin{table*}[!t]
\centering
\small
\renewcommand{\arraystretch}{0.85}
\setlength{\abovecaptionskip}{6pt}
\begin{tabular}{llcccc}
\toprule
Task & Direction & \multicolumn{2}{c}{Final best} & \multicolumn{2}{c}{Search AUC}\\
\cmidrule(lr){3-4}\cmidrule(lr){5-6}
& & LLAMBO & FGSS-BO & LLAMBO & FGSS-BO\\
\midrule
Breast RF & Accuracy $\uparrow$ & $.93187\pm.00340$ & $.93187\pm.00368$ & $.93161\pm.00365$ & $.93152\pm.00388$\\
Breast SVM & Accuracy $\uparrow$ & $.94286\pm.00197$ & $.94432\pm.00266$ & $.94282\pm.00202$ & $.94370\pm.00256$\\
Digits RF & Accuracy $\uparrow$ & $.53550\pm.25607$ & $.52727\pm.29339$ & $.46033\pm.30376$ & $.47376\pm.31506$\\
Digits SVM & Accuracy $\uparrow$ & $.99246\pm.00052$ & $.99258\pm.00036$ & $.99236\pm.00045$ & $.99235\pm.00036$\\
Diabetes RF & MSE $\downarrow$ & $.58558\pm.03102$ & $.57109\pm.04270$ & $.61024\pm.04163$ & $.61335\pm.07394$\\
Diabetes SVM & MSE $\downarrow$ & $.57768\pm.05219$ & $.51711\pm.00984$ & $.61046\pm.08684$ & $.55636\pm.05080$\\
\bottomrule
\end{tabular}
\caption{Complete six-task HPO transfer comparison: means $\pm$ sample standard deviations over six paired runs per task. Accuracy is maximized; both diabetes metrics are minimized. RF denotes RandomForest. Raw accuracy and MSE are not aggregated. LLAMBO uses the shared feasibility-preserving interface.}
\label{tab:hpo-transfer}
\end{table*}

\begin{table}[!t]
\centering
\small
\renewcommand{\arraystretch}{0.85}
\setlength{\abovecaptionskip}{6pt}
\setlength{\tabcolsep}{4pt}
\begin{tabular}{lll}
\toprule
Parameter & Domain & Scale\\
\midrule
\multicolumn{3}{l}{\textit{RandomForest}}\\
Max. depth & $[1,15]$ & Linear\\
Max. feature fraction & $[.01,.99]$ & Logit\\
Min. impurity decrease & $[0,.5]$ & Linear\\
Min. leaf sample fraction & $[.01,.49]$ & Logit\\
Min. split sample fraction & $[.01,.99]$ & Logit\\
Min. weighted leaf fraction & $[.01,.49]$ & Logit\\
\midrule
\multicolumn{3}{l}{\textit{SVM}}\\
$C$ & $[1,1000]$ & Log\\
$\gamma$ & $[.0001,.001]$ & Log\\
Tolerance & $[.00001,.1]$ & Log\\
\bottomrule
\end{tabular}
\caption{HPO domains and sampling scales inherited from LLAMBO's Bayesmark configuration, shared by both methods. Maximum depth is integer; all other parameters are continuous.}
\label{tab:hpo-spaces}
\end{table}

\begin{table}[!t]
\centering
\small
\renewcommand{\arraystretch}{0.85}
\setlength{\abovecaptionskip}{6pt}
\setlength{\tabcolsep}{3pt}
\begin{tabular}{llrr}
\toprule
Task & Metric & Gain & Bootstrap 95\% interval\\
\midrule
Breast RF & Final & $0.00000$ & $[-.00110,.00110]$\\
& AUC & $-.00009$ & $[-.00066,.00038]$\\
Breast SVM & Final & $.00147$ & $[.00037,.00293]$\\
& AUC & $.00088$ & $[-.00051,.00242]$\\
Digits RF & Final & $-.00824$ & $[-.13083,.09440]$\\
& AUC & $.01343$ & $[-.06695,.07959]$\\
Digits SVM & Final & $.00012$ & $[-.00023,.00046]$\\
& AUC & $-.00001$ & $[-.00026,.00026]$\\
Diabetes RF & Final & $.01448$ & $[-.01098,.03750]$\\
& AUC & $-.00312$ & $[-.03414,.02294]$\\
Diabetes SVM & Final & $.06057$ & $[.02708,.09707]$\\
& AUC & $.05411$ & $[.02304,.08215]$\\
\bottomrule
\end{tabular}
\caption{Paired HPO mean gains use unrounded per-seed metrics. Final denotes final best; AUC denotes search AUC. Positive favors FGSS-BO: FGSS minus LLAMBO for accuracy, reversed for MSE. Exploratory intervals are task-specific and unadjusted for multiple comparisons. LLAMBO uses the shared feasibility-preserving interface.}
\label{tab:hpo-differences}
\end{table}

\paragraph{Objective and preprocessing.}
Each configuration trains a scikit-learn model directly with five-fold cross-validation. Classification maximizes accuracy, using stratified folds; regression minimizes positive MSE. We retain the downloaded LLAMBO train/test splits and feature preprocessing. Diabetes targets are standardized once using the training split's mean and standard deviation before CV, following that runner; this is not a fold-wise target-scaling protocol. The held-out test set uses the same training-target scaling and is used only for reporting, never for candidate generation, surrogate prediction, or state updates.

RandomForest uses ten trees, random state zero, and no fixed maximum leaf count. Classification SVM uses an RBF SVC with probability estimation enabled and random state zero; regression uses an RBF SVR. Folds and training randomness are fixed. The optimization seeds index initialization configurations and candidate randomness, not independent training splits.

\paragraph{Initialization and representation.}
Both methods use the first ten public initialization configurations for their model/seed from the LLAMBO repository, canonicalized to eight decimal places without inspecting outcomes. The HPO spaces are copied from its Bayesmark configuration, with sampling scales in Table~\ref{tab:hpo-spaces}. Global and local candidate sampling respect linear, log, or logit coordinates; evidence distances remain raw range-normalized. Log-variable test steps use log-range increments. Both methods serialize candidate floats to eight decimals and outcomes to six, preventing small impurity values from collapsing under the legacy formatter. The shared interface follows the feasibility-preserving baseline, not an exact replication of free-form LLAMBO.

Both methods use the main shared LLM settings; full FGSS additionally uses the LLM reflector and bounded test policy. Feedback is true throughout, and no optional domain-prior card is enabled. CSV/JSON metrics, objective directions, budgets, and paired initialization were checked. The recorded independent audit retrained the 72 best configurations and reproduced their CV and held-out scores to floating-point precision.

Diabetes SVM reduces mean final MSE/search AUC by $10.48\%/8.86\%$; other tasks show small, tied, or mixed effects. High initialization scores constrain gains on breast RF and digits SVM. Table~\ref{tab:hpo-differences} reports all paired differences with unadjusted exploratory percentile intervals (20,000 bootstrap resamples, analysis seed 42).
\clearpage
\endgroup

\end{document}